\documentclass{article}
\usepackage[nonatbib,final]{nips_2017}

\usepackage{danudefs}
\usepackage{flexisym}
\usepackage[T1]{fontenc}
\usepackage{times}
\usepackage{url}
\usepackage{graphicx}
\usepackage{color}
\usepackage{hyperref}
\usepackage{enumerate}
\usepackage{subcaption}

\usepackage{pdflscape}
\usepackage{afterpage}
\usepackage{caption}
\usepackage{array}

\usepackage{tabularx}

\newcolumntype{H}{>{\setbox0=\hbox\bgroup}c<{\egroup}@{}}

\newcommand{\citep}{\cite}

\usepackage[font=footnotesize,labelfont=bf]{caption}

\begin{document}

\title{Learning Neural Word Salience Scores}

\author{Krasen Samardzhiev\\ University of Liverpool \\ Liverpool, UK \\ \texttt{krasensam@gmail.com} \And
 Andrew Gargett\\ Science and Technology Facilities Council \\ Daresbury, UK \\ \texttt{andrew.gargett@stfc.ac.uk} \And
 Danushka Bollegala \\ University of Liverpool \\ Liverpool, UK \\ \texttt{danushka@liverpool.ac.uk}
}

\maketitle

\begin{abstract}
Measuring the salience of a word is an essential step in numerous NLP tasks.
Heuristic approaches such as tfidf have been used so far to estimate the salience of words. 
We propose \emph{Neural Word Salience} (NWS) scores, unlike heuristics, are learnt from a corpus.
Specifically, we learn word salience scores such that, using pre-trained word embeddings as the input, can accurately predict the words that appear in a sentence, given the words that appear in the sentences preceding or succeeding that sentence.
Experimental results on sentence similarity prediction show that the learnt word salience scores perform comparably or better than some of the state-of-the-art approaches for representing sentences on benchmark datasets for sentence similarity, while using only a fraction of the training and prediction times required by prior methods.
Moreover, our NWS scores positively correlate with psycholinguistic measures such as concreteness, and imageability implying a close connection to the salience as perceived by humans.
\end{abstract}

\section{Introduction}
\label{sec:intro}

\nocite{Bollegala_ECAI_2008,Bollegala:IJCNLP:2005,Bollegala:IS:2012,Bollegala:JSAI:2007,Duc:AAAI:2011,Hernault:CICLING:2010,Hugo:EMNLP:2010,Noman:2011}

Humans can easily recognise the words that contribute to the meaning of a sentence (i.e. content words) from words that serve only a grammatical functionality (i.e. functional words).
For example, functional words such as \emph{the, an, a} etc. have limited contributions towards the overall meaning of a document and 
are often filtered out as \emph{stop words} in information retrieval systems~\citep{IR_book}.
We define the \emph{salience} $q(w)$ of a word $w$ in a given text $T$ as the semantic contribution made by $w$ towards the overall meaning of $T$. 
If we can accurately compute the salience of words, then we can develop better representations of texts that can be used in
downstream NLP tasks such as similarity measurement~\citep{Arora:ICLR:2017} or text (e.g. sentiment, entailment) classification~\citep{socher-EtAl:2011:EMNLP}.

As described later in \autoref{sec:related}, existing methods for detecting word salience can be classified into two groups:
(a) lexicon-based filtering methods such as stop word lists, or 
(b) word frequency-based heuristics such as the popular term-frequency inverse document frequency (tfidf)~\citep{tfidf} measure and its variants.
Unfortunately, two main drawbacks can be identified in common to both stop words lists and frequency-based salience scores.

First, such methods do not take into account the semantics associated with individual words when determining their salience.
For example, consider the following two adjacent sentences extracted from a newspaper article related to the visit of the Japanese Prime Minister, \emph{Shinzo Abe}, to the White House in Washington, to meet the US President \emph{Donald Trump}.
\begin{enumerate}[(a)]
\item \emph{Abe visited Washington in February and met Trump in the White House.}
\item \emph{Because the trade relations between US and Japan have been fragile after the recent comments by the US President, the Prime Minister's visit to the US can be seen as an attempt to reinforce the trade relations.}
\end{enumerate}
In Sentence (a), the Japanese person name \emph{Abe} or American person name \emph{Trump} would occur less in a corpus than the US state name \emph{Washington}. Nevertheless, for the main theme of this sentence, \emph{Japanese Prime minister met US President}, the two person names are equally important as the location they met.
Therefore, we must look into the semantics of the individual words when computing their saliences.

Second, words do not occur independently of one another in a text, and methods that compute word salience using frequency or pre-compiled stop words lists alone do not consider the contextual information.
For example, the two sentences (a) and (b) in our previous example are extracted from the same newspaper article and are adjacent.
The words in the two sentences are highly related. For example, \emph{Abe} in sentence (a) refers to the \emph{Prime Minister} in sentence (b), and \emph{Trump} in sentence (a) is refers to the \emph{US President} in sentence (b).
A human reader who reads sentence (a) before sentence (b) would expect to see some relationship between the topic discussed in (a) and that in the next sentence (b).
Unfortunately, methods that compute word salience scores considering each word independently from all other words in near by contexts, ignore such proximity relationships.

To overcome the above-mentioned disfluencies in existing word salience scores, we propose 
 an unsupervised method that first randomly initialises word salience scores, and subsequently updates them such that we can accurately predict the words in local contexts.
Specifically, we train a two-layer neural network where in the first layer we take pre-trained word embeddings of the words in a sentence $S_{i}$  as the input and compute a representation for $S_{i}$ (here onwards referred to as a \emph{sentence embedding}) as the \emph{weighted average} of the input word embeddings. The weights correspond to the word salience scores of the words in $S_{i}$.
Likewise, we apply the same approach to compute the sentence embedding for the sentence 
$S_{i-1}$ preceding $S_{i}$ and $S_{i+1}$ succeeding $S_{i}$ in a sentence-ordered corpus.
Because $S_{i-1}, S_{i}$ and $S_{i+1}$ are adjacent sentences, we would expect the sentence pairs $(S_{i}, S_{i-1})$ and $(S_{i}, S_{i+1})$ to be topically related.\footnote{$S_{i-1}$ and $S_{i+1}$ could also be topically related and produce a positive training examples in some cases. However, they are non-adjacent and possibly less related compared to adjacent sentence pairs. Because we have an abundant supply of sentences, and we want to reduce label noise in positive examples, we do not consider $(S_{i-1}, S_{i+1})$ as a positive example.}

We would expect a high degree of cosine similarity between $\vec{s}_{i}$ and $\vec{s}_{i-1}$, and $\vec{s}_{i}$ and $\vec{s}_{i+1}$,
where boldface symbols indicate vectors.
Likewise, for a randomly selected sentence $S_{j} \notin \{S_{i-1}, S_{i}, S_{i+1}\}$, 
the expect similarity between $S_{j}$ and $S_{i}$ would be low. 
We model this as a supervised similarity prediction task  and use backpropagation to update the word salience scores, keeping word embeddings fixed.
We refer to the word salience scores learnt by the proposed method as the \emph{Neural Word Salience} (NWS) scores.
We will use the contextual information of a word to learn its salience. 
However, once learnt, we consider salience as a property of a word that holds independently of its context. This enables us to use the same salience score for a word after training, without having to modify it considering the context in which it occurs.

Several remarks can be made about the proposed method for learning NWS scores.
First, we do \emph{not} require labelled data for learning NWS scores. Although we require semantically similar (positive) and semantically dissimilar (negative) pairs of sentences for learning the NWS scores, both positive and negative examples are automatically extracted from the given corpus. 
Second, we use pre-trained word embeddings as the input, and do \emph{not} learn the word embeddings as part of the learning process.
This design choice differentiates our work from previously proposed sentence embedding learning methods that jointly learn word embeddings as well as sentence embeddings~\citep{Hill:NAACL:2016,Kiros:2015,Kenter:ACL:2016}.
Moreover, it decouples the word salience score learning problem from word or sentence embedding learning problem, thereby simplifying the 
optimisation task and speeding up the learning process.
 
We use the NWS scores to compute sentence embeddings and measure the similarity between two sentences using $18$ benchmark datasets for semantic textual similarity in past SemEval tasks~\citep{Agirre:2012}. 
Experimental results show that the sentence similarity scores computed using the NWS scores and pre-trained word embeddings show a high degree of correlation with human similarity ratings in those benchmark datasets.
Moreover, we compare the NWS scores against the human ratings for psycholinguistic properties of words such as arousal, valence, dominance, imageability, and concreteness. 
Our analysis shows that NWS scores demonstrate a moderate level of correlation with concreteness and imageability ratings, despite not being specifically trained to predict such psycholinguistic properties of words.

\section{Related Work}
\label{sec:related}

Word salience scores have long been studied in the information retrieval community~\citep{IR_book}. 
Given a user query described in terms of one or more keywords, an information retrieval system must find the most relevant documents to the user query from a potentially large collection of documents.
Word salience scores based on term frequency, document frequency, and document length have been proposed such as
tfidf and BM25~\citep{Robertson:1997}.

Our proposed method learns word salience scores by creating sentence embeddings. 
Next, we briefly review such sentence embedding methods and explain the differences between the sentence embedding learning problem and
word salience learning problem.

Sentences have a syntactic structure and the ordering of words affects the meaning expressed in the sentence.
Consequently, compositional approaches for computing sentence-level semantic representations from word-level semantic representations have used numerous linear algebraic operators such as vector addition, element-wise multiplication, multiplying by a matrix or a tensor~\citep{blacoe-lapata:2012:EMNLP-CoNLL,Mitchell:ACL:2008}.

Alternatively to applying nonparametric operators on word embeddings to create sentence embeddings,
recurrent neural networks can learn the optimal weight matrix that can produce an accurate sentence embedding when
repeatedly applied to the constituent word embeddings.
For example, skip-thought vectors~\citep{Kiros:2015} use bi-directional LSTMs to predict the words in the order they appear in the previous and next sentences given the current sentence. 
Although skip-thought vectors have shown superior performances in supervised tasks, its performance on unsupervised tasks has been sub-optimal~\citep{Arora:ICLR:2017}.
Moreover, training bi-directional LSTMs from large datasets is time consuming and we also need to perform LSTM inference in order to create the embedding for unseen sentences at test time, which is time consuming compared to weighted addition of the input word embeddings.
FastSent~\citep{Hill:NAACL:2016} was proposed as an alternative lightweight approach for sentence embedding where a softmax objective is optimised to predict the occurrences of words in the next and the previous sentences, ignoring the ordering of the words in the sentence.

Surprisingly, averaging word embeddings to create sentence embeddings has shown comparable performances to sentence embeddings that are learnt using more sophisticated word-order sensitive methods.
For example, \cite{Arora:ICLR:2017} proposed a method to find the optimal weights for combining word embeddings when creating sentence embeddings using unigram probabilities, by maximising the likelihood of the occurrences of words in a corpus.
Siamese CBOW~\citep{Kenter:ACL:2016} learns word embeddings such that we can accurately compute sentence embeddings by averaging the word embeddings.
Although averaging is an order insensitive operator, \cite{Adi:2016} empirically showed that it can accurately predict the content and word order in sentences.
This can be understood intuitively by recalling that words that appear between two words are often different in contexts where those two words are swapped.
For example, in the two sentences ``\emph{Ostrich} is a large \emph{bird} that lives in Africa'' and ``Large \emph{birds} such as \emph{Ostriches} live in Africa'',
the words that appear in between \emph{ostrich} and \emph{bird} are different, giving rise to different sentence embeddings even when sentence embeddings are
computed by averaging the individual word embeddings.
Instead of considering all words equally for sentence embedding purposes, attention-based models~\citep{Hahn:EMNLP:2016,Yin:TACL:2016,CSE:2016} learn the amount of weight (attention) we must assign to each word in a given context.

Our proposed method for learning NWS scores are based on the prior observation that averaging is an effective heuristic for creating sentence embeddings from word embeddings.
However, unlike sentence embedding learning methods that do not learn word salience scores~\citep{he-lin:2016:N16-1,Yin:TACL:2016} , our goal in this paper is to learn word salience scores and not sentence embeddings.
We compute sentence embeddings only for the purpose of evaluating the word salience scores we learn.
Moreover, our work differs from Siamese CBOW~\citep{Kenter:ACL:2016} in that we do not learn word embeddings but take pre-trained word embeddings as the input for learning word salience scores.
NWS scores we learn in this paper are also different from the salience scores learnt by \cite{Arora:ICLR:2017} because they do not constrain their word salience scores such that they can be used to predict the words that occur in adjacent sentences.

\section{Neural Word Salience Scores}

\begin{figure}[t]
\centering
\includegraphics[width=8cm]{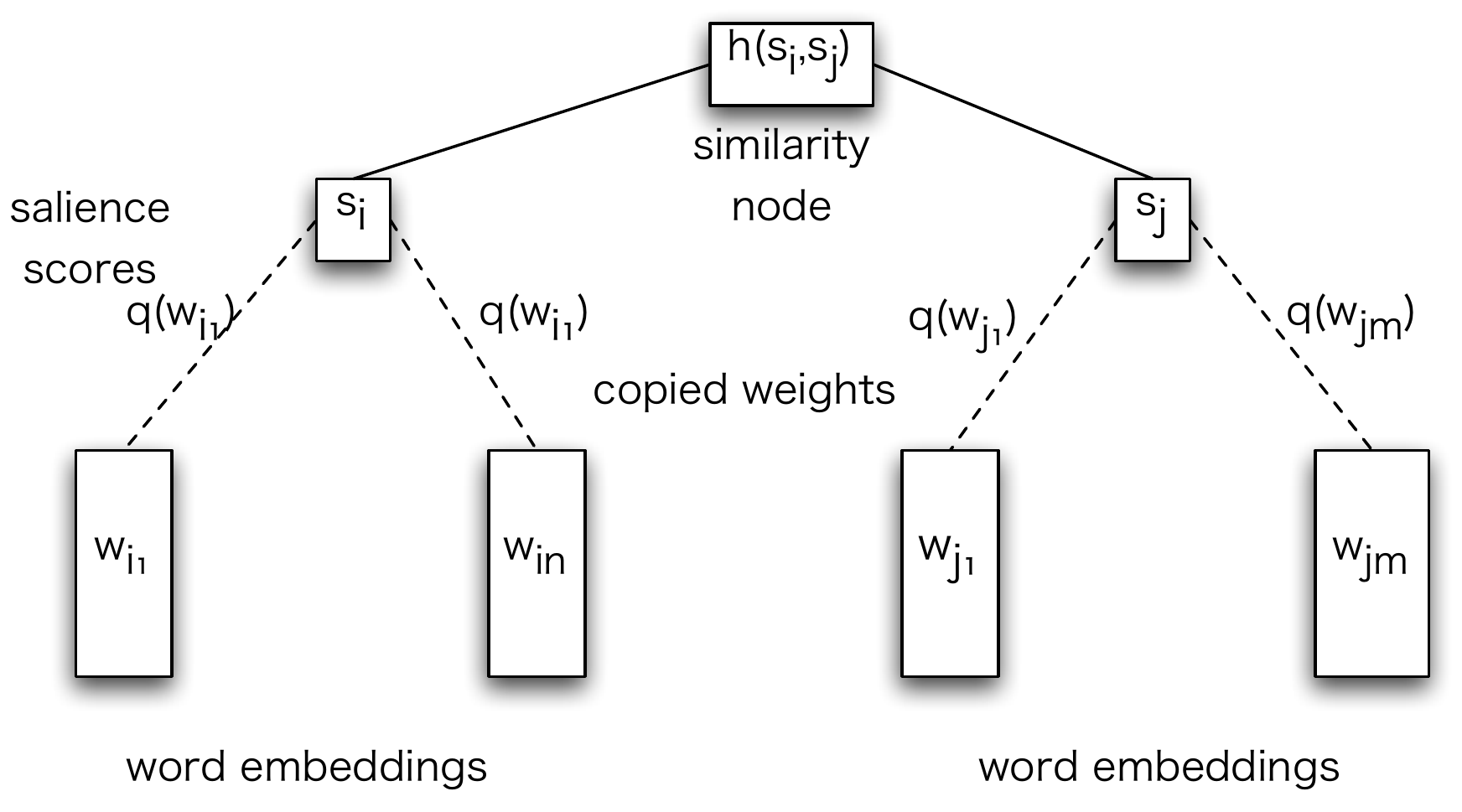}
\caption{Overview of the proposed neural word salience learning method. Given two sentences $(S_{i}, S_{j})$, we learn the salience scores of words $q(w)$ such that we can predict the similarity between the two sentences using their embeddings $\vec{s}_{i}$, $\vec{s}_{j}$. Difference between predicted similarity and actual label is considered as the error and its gradient is backpropagated through the network to update $q(w)$.}
\label{fig:model}
\end{figure}

Let us consider a vocabulary $\cV$ of words $w \in \cV$. For the simplicity of exposition, we limit the vocabulary to unigrams
but note that the proposed method can be used to learn salience scores for arbitrary length $n$-grams. 
We assume that we are given $d$-dimensional pre-trained word embeddings $\vec{w} \in \R^{d}$ for the words in $\cV$.
Let us denote the NWS score of $w$ by $q(w) \in R$.
We learn $q(w)$  such that the similarity between two adjacent sentences $\cS_{i}$ and $\cS_{i-1}$, or $\cS_{i}$ and $\cS_{i+1}$ in a 
sentence-ordered corpus $\cC$ is larger than that between two non-adjacent sentences $\cS_{i}$ and $\cS_{j}$, where $j \notin \{i-1, i, i+1\}$.
Let us further represent the two sentence $\cS_{i} = \{w_{i1}, \ldots, w_{in}\}$ and $\cS_{j} = \{w_{j1}, \ldots, w_{jm}\}$ by the sets of words in those sentences.
Here, we assume the corpus to contain sequences of ordered sentence such as in a newspaper article, a book chapter or a blog post.

The neural network we use for learning $q(w)$ is shown in \autoref{fig:model}. 
The first layer computes the embedding of a sentence $\cS$, $\vec{s} \in \R^{d}$ using \autoref{eq:sentemb}, which is the weighted-average of the individual word embeddings.
\begin{equation}
\label{eq:sentemb}
\vec{s} = \sum_{w \in \cS} q(w) \vec{w}
\end{equation}
We use \eqref{eq:sentemb} to compute embeddings for two sentences $\cS_{i}$ and $\cS_{j}$ denoted respectively by $\vec{s}_{i}$ and $\vec{s}_{j}$.
Here, the same set of salience scores $q(w)$ are used for computing both $\vec{s}_{i}$ and $\vec{s}_{j}$, which resembles a Siamese neural network architecture.

The root node computes the similarity $h(\vec{s}_{i}, \vec{s}_{j})$ between two sentence embeddings. Different similarity (alternatively dissimilarity or divergence) functions such as cosine similarity, $\ell_{1}$ distance, $\ell_{2}$ distance, Jenson-Shannon divergence etc. can be used as $h$.
As a concrete example, here we use softmax of the inner-products as follows:
\begin{equation}
\small
\label{eq:softmax}
h(\vec{s}_{i}, \vec{s}_{j}) = \frac{\exp \left( \vec{s}_{i}\T\vec{s}_{j} \right) }{\sum_{\cS_{k} \in \cC} \exp \left( \vec{s}_{i}\T\vec{s}_{k} \right)}
\end{equation} 
Ideally, the normalisation term in the denominator in the softmax must be taken over all the sentences $\cS_{k}$ in the corpus~\citep{andreas-klein:2015:NAACL-HLT}.
However, this is computationally expensive in most cases except for extremely small corpora.
Therefore, following noise-contrastive estimation~\citep{Gutmann:2012}, we approximate the normalisation term using a randomly sampled set of $K$ sentences, where $K$ is typically less than $10$.
Because the similarity between two randomly sampled sentences is likely to be smaller than, for example, two adjacent sentences, we can see this sampling process as randomly sampling \emph{negative} training instances from the corpus.

For two sentences $\cS_{i}$ and $\cS_{j}$ we consider them to be similar (positive training instance) if $j \in \{i - 1, i + 1\}$, and denote this by the target label $t = 1$. On the other hand, if the two sentences are non-adjacent (i.e. $j \notin \{i - 1, i + 1\}$), then we consider the pair $(\cS_{i}, \cS_{j})$ to form a negative training instance, and denote this by $t = 0$.\footnote{It is possible in theory that two non-adjacent sentences could be similar, but the likelihood of this event is small and can be safely ignored in practice.}
This assumption enables us to use a sentence-ordered corpus for selecting both positive and negative training instances required for learning NWS scores.

Using $t$ and $h(\vec{s}_{i}, \vec{s}_{j})$ above, we compute the cross-entropy error $E(t, (\cS_{i}, \cS_{j}))$ for an instance $(t, (\cS_{i}, \cS_{j}))$ as follows:
\begin{equation}
\small
 \label{eq:error}
 E(t, (\cS_{i}, \cS_{j})) = t \log \left( h(\vec{s}_{i}, \vec{s}_{j}) \right) + (1-t) \log \left( 1 - h(\vec{s}_{i}, \vec{s}_{j}) \right)
\end{equation}
Next, we backpropagate the error gradients via the network to compute the updates as follows:
\begin{equation}
\small
 \label{eq:g1}
 \frac{\partial E}{\partial q(w)} = \frac{\left( t - h(\vec{s}_{i}, \vec{s}_{j}) \right)}{h(\vec{s}_{i}, \vec{s}_{j}) (1 - h(\vec{s}_{i}, \vec{s}_{j}))} \frac{\partial h(\vec{s}_{i}, \vec{s}_{j})}{\partial q(w)}
\end{equation}
Here, we drop the arguments of the error and simply write it as $E$ to simplify the notation.
To compute $\frac{\partial h(\vec{s}_{i}, \vec{s}_{j})}{\partial q(w)}$ let us define
\begin{equation}
\small
 \label{eq:g}
 g(\vec{s}_{i}, \vec{s}_{j}) = \log \left( h(\vec{s}_{i}, \vec{s}_{j}) \right)
\end{equation}
From which we have, 
\begin{equation}
\small
\label{eq:grad-h}
\frac{\partial h(\vec{s}_{i}, \vec{s}_{j})}{\partial q(w)} = h(\vec{s}_{i}, \vec{s}_{j}) \frac{\partial g(\vec{s}_{i}, \vec{s}_{j})}{\partial q(w)} .
\end{equation}

We can then compute $\frac{\partial g}{\partial q(w)}$ as follows:
{\small
\begin{align}
\label{eq:grad-g}
& \cI[w \in \cS_{i}] \vec{w}\T\vec{s}_{j} + \cI[w \in \cS_{j}] \vec{w}\T\vec{s}_{i} \\
 &- \log \left( \sum_{k} \exp \left( \vec{s}_{i}\T\vec{s}_{j} \right) \left( \cI[w \in \cS_{i}] \vec{w}\T\vec{s}_{k} + \cI[w \in \cS_{k}] \vec{w}\T\vec{s}_{i} \right) \right)
\end{align}}
Here, the indicator function $\cI$ is given by \eqref{eq:I}.
\begin{equation}
\small
\label{eq:I}
\cI[\theta] = \begin{cases} 1 & \theta \text{ is True} \\ 0 & \text{otherwise} \end{cases}
\end{equation}

Substituting \eqref{eq:I}, \eqref{eq:grad-g}, in \eqref{eq:g1} we compute $\frac{\partial E}{\partial q(w)}$ and use stochastic gradient descent 
with initial learning rate set to $0.01$ and subsequently scheduled by AdaGrad~\cite{Duchi:JMLR:2011}.
The NWS scores are randomly initialised in our experiments.

\section{Experiments}
\label{sec:exp}

We use the Toronto books corpus\footnote{\url{http://yknzhu.wixsite.com/mbweb}} as our training dataset. 
This corpus contains 81 million sentences from 11,038 books, and has been used
as a training dataset in several prior work on sentence embedding learning. 
We convert all sentences to lowercase and tokenise using the Python NLTK\footnote{\url{http://www.nltk.org/}} punctuation tokeniser. 
No further pre-processing is conduced beyond tokenisation.
The proposed method is implemented using TensorFlow\footnote{\url{https://www.tensorflow.org/}} and
executed on a NVIDIA Tesla K40c 2880 GPU. 
The source code is submitted as a supplementary and will be publicly released upon paper acceptance.

\subsection{Measuring Semantic Textual Similarity}
\label{sec:sts}

It is difficult to evaluate the accuracy of word salience scores by direct manual inspection. 
Moreover, there does not exist any datasets where human annotators have manually rated words for their salience.
Therefore, we resort to extrinsic evaluations, where, we first use \eqref{eq:sentemb} to create the sentence embedding
 for a given sentence using pre-trained word embeddings and the NWS scores computed using the proposed method.
Next, we measure the semantic textual similarity (STS) between two sentences by the cosine similarity between the corresponding sentence embeddings.
Finally, we compute the correlation between human similarity ratings for sentence pairs in benchmark datasets for STS and the similarity scores
computed following the above-mentioned procedure.
If there exists a high degree of correlation between the sentence similarity scores computed using the NWS scores and human ratings, 
then it can be considered as empirical support for the accuracy of the NWS scores.
As shown in \autoref{tbl:main}, we use 18 benchmark datasets from SemEval STS tasks from years 2012~\citep{Agirre:2012}, 2013~\citep{Agirre:2013}, 2014~\citep{Agirre:SemEval:2014}, and 2015~\citep{SemEval:Task2:Agirre:2015}. 
Note that the tasks with the same name in different years actually represent different tasks.

We use Pearson correlation coefficient as the evaluation measure. For a list of $n$ ordered pairs of ratings $\{(x_{i}, y_{i})\}_{i=1}^{n}$,
the Pearson correlation coefficient between the two ratings, $r(\vec{x}, \vec{y})$, is computed as follows:
\begin{equation}
 \label{eq:Pearson}
 r(\vec{x}, \vec{y}) = \frac{\sum_{i=1}^{n}(x_{i} - \bar{x})(y_{i} - \bar{y})}{\sqrt{\sum_{i=1}^{n}(x_{i} - \bar{x})^{2}} \sqrt{\sum_{i=1}^{n}(y_{i} - \bar{y})^{2}}}
\end{equation}
Here, $\bar{x} = \frac{1}{n}\sum_{i=1}^{n} x_{i}$ and $\bar{y} = \frac{1}{n}\sum_{i=1}^{n} y_{i}$. 
Pearson correlation coefficient is invariant against linear transformations of the similarity scores, which is suitable for comparing similarity scores assigned to the same set of items by two different methods (human ratings vs. system ratings).

We use the Fisher transformation~\citep{Fisher:1915} to test for the statistical significance of Pearson correlation coefficients.
Fisher transformation, $F(r)$, of the Pearson correlation coefficient $r$ is given by \eqref{eq:ft}.
\begin{equation}
 \label{eq:ft}
 F(r) = \frac{1}{2} \log \left( \frac{1+r}{1-r} \right)
\end{equation}
Then, $95\%$ confidence intervals are given by \eqref{eq:z}.
\begin{equation}
 \label{eq:z}
\tanh \left(F(r) \pm \frac{1.96}{\sqrt{n - 3}} \right)
\end{equation}

We consider two baseline methods in our evaluations as described next.
\begin{description}

\item[Averaged Word Embeddings (AVG)]
As a baseline that does not use any salience scores for words when computing sentence embeddings, we use \emph{Averaged Word Embeddings} (AVG)
where we simply add all the word embeddings of the words in a sentence and divide from the total number of words to create a sentence embedding.
This baseline demonstrates the level of performance we would obtain if we did not perform any word salience-based weighting in \eqref{eq:sentemb}.

\item[Inverse Sentence Frequency (ISF)]
As described earlier in \autoref{sec:related}, term frequency is not a useful measure for discriminating salient vs. non-salient words in short-texts because
it is rare for a particular word to occur multiple times in a short text such as a sentence.
However, (inverse of) the number of different sentences in which a particular word occurs is a useful method for identifying salient features
because non-content stop words are likely to occur in any sentence, irrespective of the semantic contribution to the topic of the sentence.
 Following the success of Inverse Document Frequency (IDF) in filtering out high frequent words in text classification tasks~\citep{Joachims:ECML:98},
we define \emph{Inverse Sentence Frequency} (ISF) of a word as the reciprocal of the number of sentences in which that word appears in a corpus.
Specifically, ISF is computed as follows:
\begin{equation}
\label{eq:idf}
\textrm{ISF}(w) = \log \left( 1 + \frac{\mbox{total no. of sentences in the corpus}}{\mbox{no. of sentences containing $w$}} \right)
\end{equation}
\end{description}

In \autoref{tbl:main}, we compare NWS against AVG, ISF baselines. 
\textbf{SMOOTH} is the unigram probability-based smoothing method proposed by \cite{Arora:ICLR:2017}.\footnote{Corresponds to the
GloVe-W method in the original publication.}
We compute sentence embeddings for NWS, AVG and ISF using pre-trained 300 dimensional GloVe embeddings
trained from the Toronto books corpus using contextual windows of 10 tokens.\footnote{We use the GloVe implementation by the original authors available at \url{https://nlp.stanford.edu/projects/glove/}}
For reference purposes we show the level of performance we would obtain
if we had used sentence embedding methods such as, skip-thought~\citep{Kiros:2015}, and Siamese-CBOW~\citep{Kenter:ACL:2016}. 
Note that however, sentence embedding methods do not necessarily compute word salience scores.
For skip-thought, Siamese~CBOW and SMOOTH methods we report the published results in the original papers.
Because \cite{Kiros:2015} did not report results for skip-thought on all 18 benchmark datasets used here, we report
the re-evaluation of skip-thought on all 18 benchmark datasets by \cite{Wieting:ICLR:2016}.


\begin{table*}[t]
\centering
\caption{Performance on STS benchmarks.}
\label{tbl:main}
\begin{tabular}{| l || c | H  c| c | c | c | c |}\hline 
Dataset & SMOOTH & Glove-WR & skip-thought & Siamese-CBOW & AVG & ISF & NWS \\ \hline \hline
\underline{2012} & & & & & & & \\
MSRpar  & 43.6 & 35.6 & 5.6 & $\mathbf{43.8}$ & 28.4 & 39.1 & 28.5 \\
OnWN  & 54.3 & $\mathbf{66.2}^{*}$ & 60.5 & 64.4 & 47.1 & 60.5 & $\mathbf{65.5}^{*}$ \\
SMTeuroparl & $\mathbf{51.1}^{*}$ & 49.9 & 42.0 & 45.0 & 37.1 & 44.5 & $50.1$ \\
SMTnews  & 42.2 &  $45.6^{*}$ & 39.1 & 39.0 & 32.2 & 34.9 & $\mathbf{44.7}^{*}$ \\ \hline
\underline{2013} & & & & & & &   \\
FNWN  & 23.0 & $\mathbf{39.4}$ & $\mathbf{31.2}$ & 23.2 & 26.9 & 29.4 & 25.2 \\
OnWN  & $68.0^{*}$ & $\mathbf{82.8}^{*}$ & 24.2 & 49.9 & 25.0 & 63.2 & $\mathbf{78.1}^{*}$ \\
headlines  & 63.8 & $\mathbf{69.2}^{*}$ & 38.6 & $\mathbf{65.3}^{*}$ & 40.2 & 59.4 & 57.0 \\ \hline
\underline{2014} & & & & & & &  \\
OnWN  & 68.0 & $\mathbf{82.8}^{*}$ & 46.8 & 60.7 & 41.1 & 68.5 & $\mathbf{80.8}^{*}$ \\
deft-forum  & 29.1 & $\mathbf{41.2}$ & 37.4 & $\mathbf{40.8}$ & 27.1 & 37.1 & 29.9 \\
deft-news  & $\mathbf{68.5}$ & $\mathbf{69.4}$ & 46.2 & 59.1 & 48.8 & 63.6 & 65.4 \\
headlines  & 59.3 & $\mathbf{64.7}^{*}$ & 40.3 & $\mathbf{63.6}^{*}$ & 41.9 & 58.8 & 56.2 \\
images  & $74.1^{*}$ & $\mathbf{82.6}^{*}$ & 42.6 & 65.0 & 35.3 & 66.3 & $\mathbf{75.9}^{*}$ \\
tweet-news  & 57.3 & $70.1^{*}$ & 51.4 & $\mathbf{73.2}^{*}$ & 41.7 & 57.1 & $64.5^{*}$ \\ \hline
\underline{2015} & & & & & & & \\
answers-forums  & 41.4 & $\mathbf{63.9}^{*}$ & 27.8 & 21.8 & 25.7 & 37.6 & $\mathbf{49.6}^{*}$ \\
answers-students  & 61.5 & $\mathbf{70.4}$ & 26.6 & 36.7 & 56.5 & 67.1 & $\mathbf{68.0}$ \\
belief  & 47.7 & $\mathbf{71.8}^{*}$ & 45.8 & 47.7 & 29.3 & 43.2 & $\mathbf{54.3}^{*}$ \\
headlines  & 64.0 & $\mathbf{70.7}^{*}$ & 12.5 & 21.5 & 49.3 & $\mathbf{65.4}$ & 65.3 \\
images  & $75.4^{*}$ & $\mathbf{81.5}^{*}$ & 21 & 25.6 & 49.8 & 66.1 & $\mathbf{76.6}^{*}$ \\ \hline \hline
Overall Average & 55.1 & $\mathbf{64.3}^{*}$ & 35.5 & 47.0 & 38.0 & 53.4 & $\mathbf{57.6}$ \\
\hline
\end{tabular}
\end{table*}

Statistically significant improvements over the ISF baseline are indicated by an asterisk $*$, 
whereas the best results on each benchmark dataset are shown in bold. 
From \autoref{tbl:main}, we see that between the two baselines AVG and ISF, ISF consistently outperforms AVG in all benchmark datasets.
In 9 out of the 18 benchmarks, the proposed NWS scores report the best performance.
Moreover, in 9 datasets NWS statistically significantly outperforms the ISF baseline.
Siamese-CBOW reports the best results in 5 datasets, whereas SMOOTH reports the best results in 2 datasets.
Overall, NWS stands out as the best performing method among the methods compared in \autoref{tbl:main}.

\begin{table}[t]
\centering
\caption{Effect of word embeddings.}
\label{tbl:prop}
\begin{tabular}{|l||H  H H H H H c  c  c  H H H|}\hline
Dataset & \multicolumn{3}{H}{AVG} & \multicolumn{3}{H}{ISF} & \multicolumn{3}{c}{NWS with pre-trained} & \multicolumn{3}{H|}{NWS (RND)} \\ 
&  SGNS & CBOW & GloVe & SGNS & CBOW & GloVe & SGNS & CBOW & GloVe & SGNS & CBOW & GloVe \\  \hline \hline
\underline{2012} & & & & & & & & & & & &  \\
MSRpar  & \textbf{31.35} & 28.41 & 28.45 & 25.87 & 24.21 & \textbf{39.06} & 14.27 & 24.15 & \textbf{28.47} & 17.14 & 21.77 & 20.74 \\
OnWN  & 57.55 & \textbf{66.36} & 47.11 & 64.67 & \textbf{67.11} & 60.52 & 59.76 & 61.25 & \textbf{65.50} & 48.37 & 55.01 & 55.52 \\
SMTeuroparl & 45.16 & 43.77 & 37.07 & 42.15 & 42.67 & 44.54 & 41.04 & 45.51 & \textbf{50.12} & 26.55 & 26.19 & 31.96 \\
SMTnews  & 40.4 & 42.2 & 32.22 & 47.34 & \textbf{48.34} & 34.99 & 43.42 & \textbf{46.94} & 44.73 & 23.72 & 25.14 & 30.4 \\ \hline
\underline{2013} & & & & & & & & & & & &  \\
FNWN  & 41.6 & \textbf{40.82} & 26.98 & 39.85 & \textbf{44.68} & 29.37 & 21.47 & \textbf{29.31} & 25.21 & 13.14 & 15.55 & 23.91 \\
OnWN  & 53.83 & 63.81 & 25.02 & 74.77 & \textbf{79.87} & 63.28 & 67.37 & 70.04 & \textbf{78.06} & 68.11 & 72.75 & 74.65 \\
headlines  & 60.86 & \textbf{63.47} & 40.25 & 64.62 & \textbf{65.09} & 59.44 & 57.05 & \textbf{57.46} & 57.02 & 35.86 & 37.14 & 35.51 \\ \hline
\underline{2014} & & & & & & & & & & & &  \\
OnWN  & 63.87 & 71.25 & 41.13 & 77.46 & \textbf{81.34} & 68.53 & 73.06 & 73.71 & \textbf{80.83} & 65.09 & 72.58 & 73.18 \\
deft-forum  & 36.69 & \textbf{43.38} & 27.13 & 38.83 & 43.64 & 37.08 & 28.62 & \textbf{32.49} & 29.90 & 17.60 & 16.98 & 23.79 \\
deft-news  & 64.53 & \textbf{65.41} & 48.79 & 68.6 & 67.97 & 63.63 & 59.63 & 61.95 & \textbf{65.35} & 41.51 & 33.90 & 49.61 \\
headlines  & 59.52 & \textbf{61.93} & 41.92 & 63.86 & 63.75 & 58.81 & 56.05 & 55.64 & \textbf{56.20} & 36.79 & 41.23 & 34.11 \\
images  & 70.29 & 75.46 & 35.28 & 76.85 & 80.52 & 66.26 & 76.94 & \textbf{78.08} & 75.88 & 60.35 & 70.61 & 66.66 \\
tweet-news  & 46.27 & 63.09 & 41.67 & 57.19 & 66.61 & 57.11 & 61.49 & \textbf{66.41} & 64.46 & 49.60 & 49.20 & 52.18 \\ \hline
\underline{2015} & & & & & & & & & & & &  \\
answers-forums  & 37.29 & 48.16 & 25.7 & 50.92 & 60.97 & 37.62 & 36.35 & 46.78 & \textbf{49.65} & 24.21 & 25.17 & 33.43 \\
answers-students  & 67.66 & \textbf{72.28} & 56.48 & 69.07 & 69.24 & 67.1 & 59.53 & 59.92 & \textbf{68.01} & 56.88 & 43.81 & 60.45 \\
belief  & 39.09 & \textbf{56.18} & 29.27 & 51.7 & 65.34 & 43.15 & 51.97 & \textbf{55.65} & 54.27 & 32.07 & 33.33 & 43.99 \\
headlines  & 65.00 & \textbf{67.59} & 49.34 & 69.4 & 70.19 & 65.49 & 61.24 & 63.04 & \textbf{65.32} & 43.77 & 48.41 & 43.28 \\
images  & 72.77 & 77.5 & 49.81 & 77.02 & 81.74 & 66.1 & 77.67 & \textbf{78.39} & 76.55 & 54.7 & 67.97 & 65.37 \\ \hline \hline
Overall Average & 52.98 & \textbf{58.39} & 37.97 & \textbf{58.89} & \textbf{62.40} & 53.44 & 52.60 & 55.92 & \textbf{57.52} & 39.74 & 42.04 & 45.48 \\ \hline 
\end{tabular}
\end{table}

Our proposed method for learning NWS scores does not assume any specific properties of a particular word embedding learning algorithm.
Therefore, in principle, we can learn NWS scores using \emph{any} pre-trained set of word embeddings.
To evaluate the accuracy of the word salience scores computed using different word embeddings, we conduct the following experiment.
We use SGNS, CBOW and GloVe word embedding learning algorithms to learn 300 dimensional word embeddings from the Toronto books corpus.\footnote{We use the implementation of word2vec from \url{https://github.com/dav/word2vec}}
The vocabulary size, cut-off frequency for selecting words, context window size are are kept fixed across different word embedding learning methods for the consistency of the evaluation. We then trained NWS with each set of word embeddings.
Performance on STS benchmarks is shown in \autoref{tbl:prop}, where the best performance is bolded.

From \autoref{tbl:prop}, we see that GloVe is the best among the three word embedding learning methods compared in \autoref{tbl:prop} for producing pre-trained word embeddings for the purpose of learning NWS scores.
In particular, NWS scores reports best results with GloVe embeddings in 10 out of the 18 benchmark datasets, whereas with CBOW embeddings it obtains the best results in the remaining 8 benchmark datasets.

\begin{figure}[t!]
\begin{subfigure}[t]{0.5\textwidth}
\centering
\includegraphics[width=80mm]{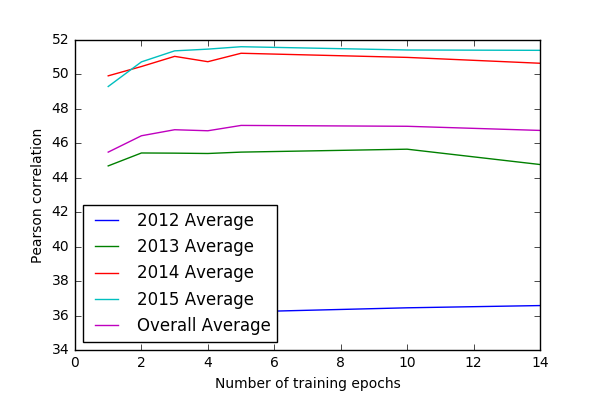}
\caption{GloVe}
\label{fig:glove}
\end{subfigure}
\hfill
\begin{subfigure}[t]{0.5\textwidth}
\centering
\includegraphics[width=80mm]{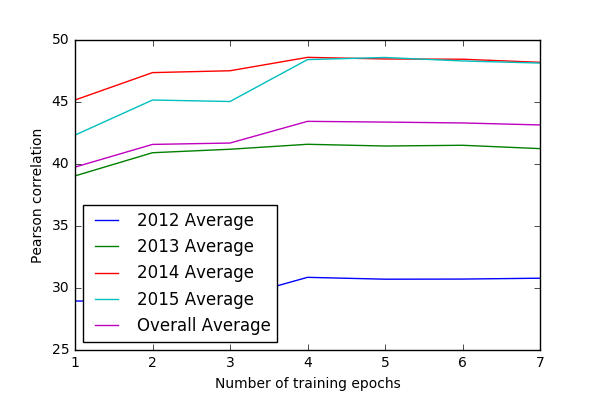}
\caption{SGNS}
\label{fig:sgns}
\end{subfigure}
\caption{Pearson correlations on STS benchmarks against the number of training epochs}
\label{fig:epochs}
\end{figure}

Figures~\ref{fig:glove} and \ref{fig:sgns} show the Pearson correlation coefficients on STS benchmarks obtained by NWS scores computed 
respectively for GloVe and SGNS embeddings. 
We plot training curves for the average correlation over each year's benchmarks as well as the overall average over the 18 benchmarks. 
We see that for both embeddings the training saturates after about five or six epochs.
This ability to learn quickly with a small number of epochs is attractive because it reduces the training time.

\subsection{Correlation with Psycholinguistic Scores}
\label{sec:psycho}

\begin{table}[t!]
\centering
\caption{Pearson correlation coefficients against Psycholinguistic ratings of words in the ANEW and MRC databases.}
\label{tbl:psycho}
\begin{tabular}{|l||p{8mm}|p{8mm}|p{8mm}|p{8mm}|c|}\hline
Embeddings & Arousal & Conc. & Dom. & Img. & Valance \\ \hline \hline
GloVe & 0.03 & 0.26 & 0.09 & 0.25 & 0.03 \\
CBOW & 0.04 & -0.35 & -0.04 & -0.37 & 0.04 \\
SGNS & -0.01 & 0.27 & 0.06 & 0.27 & -0.01 \\ \hline
\end{tabular}
\end{table}

Prior work in psycholinguistics show that there is a close connection between the emotions felt by humans and the words they read in a text.
\emph{Valence} (the pleasantness of the stimulus), \emph{arousal} (the intensity of emotion provoked by the stimulus), and \emph{dominance} (the degree of control exerted by the stimulus) contribute to how the meanings of words affect human psychology, and often referred to as the \emph{affective} meanings of words. 
\cite{Mandera:2015} show that by using SGNS embeddings as features in a $k$-Nearest Neighbour classifier, it is possible to accurately extrapolate the affective meanings of words. 
Moreover, perceived psycholinguistic properties of words such as \emph{concreteness} (how ``palpable'' the object the word refers to) and \emph{imageability} (the intensity with which a word arouses images) have been successfully predicted using 
word embeddings~\citep{Turney:EMNLP:2011,Paetzold:NAACL:2016}. 
For example, \cite{Turney:EMNLP:2011} used the cosine similarity between word embeddings obtained via Latent Semantic Analysis (LSA)~\citep{LSA} to predict the concreteness and imageability ratings of words. 

On the other hand, prior work studying the relationship between human reading patterns using eye-tracking devices show that there exist a high positive correlation between word salience and reading times~\citep{Dziemianko:2013,Hahn:EMNLP:2016}. 
For example, humans pay more attention to words that carry meaning as indicated by the longer fixation times.
Therefore, an interesting open question is that \emph{what psycholinguistic properties of words, if any, are related to the NWS scores we learn in a purely unsupervised manner from a large corpus?}
To answer this question empirically, we conduct the following experiment.
We used the Affected Norms for English Words (ANEW) dataset created by Warriner et al.~\cite{Warriner:2013}, which contains valence, arousal, and dominance ratings collected via crowd sourcing for 13,915 words.
Moreover, we obtained concreteness and imageability ratings for 3364 words from the MRC psycholinguistic database.
We then measure the Pearson correlation coefficient between NWS scores and each of the psycholinguistic ratings as shown in \autoref{tbl:psycho}.

We see a certain degree of correlation between NWS scores computed for all three word embeddings and the  concreteness scores.
Both GloVe and SGNS show moderate positive correlations for concreteness, whereas CBOW shows a moderate negative correlation for the same.
A similar trend can be observed for imageability ratings in \autoref{tbl:psycho}, where GloVe and SGNS correlates positively with imageability, while
CBOW correlates negatively. 
Moreover, no correlation could be observed for arousal, valance and dominance ratings.
This result shows that NWS scores are not correlated with affective meanings of words (arousal, dominance, and valance), but show a moderate level of correlation with perceived meaning scores (concreteness and imageability).

\section{Conclusion}

We proposed a method for learning Neural Word Salience scores from a sentence-ordered corpus, without requiring any manual data annotations.
To evaluate the learnt salience scores, we computed sentence embeddings as the linearly weighted sum over pre-trained word embeddings.
Our experimental results show that the proposed NWS scores outperform baseline methods, previously proposed word salience scores and
sentence embedding methods on a range of benchmark datasets selected from past SemEval STS tasks.
Moreover, the NWS scores shows interesting correlations with perceived meaning of words indicated by concreteness and imageability psycholinguistic ratings.

\bibliographystyle{abbrvurl}
\bibliography{salience.bib}
\end{document}